\documentclass{article}
\usepackage{spconf,amsmath,graphicx,hyperref}

\usepackage{amssymb}
\usepackage{amsfonts}
\usepackage{mathtools}
\usepackage{bbm}

\usepackage{cite}
\usepackage{url}
\usepackage{hyperref}

\usepackage{cleveref} %

\usepackage{caption}
\usepackage{float}

\usepackage{xcolor}

\usepackage{times}
\usepackage{epsfig}
\usepackage{graphicx}
\usepackage{amsmath}
\usepackage{amssymb}
\usepackage{url}
\usepackage[numbers,sort&compress]{natbib}
\usepackage{graphicx}
\usepackage{booktabs}
\usepackage{float}
\usepackage{multirow}
\usepackage{makecell}

\crefname{section}{Sec.}{Secs.}
\Crefname{section}{Section}{Sections}
\Crefname{table}{Table}{Tables}
\crefname{table}{Tab.}{Tabs.}
\Crefname{figure}{Figure}{Figures}
\crefname{figure}{Fig.}{Figs.}

\definecolor{firstplace}{RGB}{255,179,179}   
\definecolor{secondplace}{RGB}{255,217,179} 
\definecolor{thirdplace}{RGB}{255,255,180}   

\title{Multimodal-Prior-Guided Importance Sampling for Hierarchical Gaussian Splatting in Sparse-View Novel View Synthesis}
\name{ Kaiqiang Xiong\textsuperscript{1,2}, Zhanke Wang\textsuperscript{1}, Ronggang Wang\textsuperscript{1,2,3 *}\thanks{ * Ronggang Wang is the corresponding author(rgwang@pkusz.edu.cn). } \thanks{This work is financially supported by Guangdong Provincial Key Laboratory of Ultra High Definition Immersive Media Technology(Grant No. 2024B1212010006), this work is also financially supported for Outstanding Talents Training Fund in Shenzhen, Shenzhen Science and Technology Program(Grant No. SYSPG20241211173440004,   and Grant No. RCJC20200714114435057), R24115SG MIGU-PKU META VISION TECHNOLOGY INNOVATION LAB. This work is also supported by the Key Research and Development Program of Pengcheng Laboratory under Grant PCL2024A02.} }

\address{{\textsuperscript{1}Guangdong Provincial Key Laboratory of Ultra High Definition Immersive Media Technology,} \\
{Shenzhen Graduate School, Peking University, China} \\
{\textsuperscript{2}Peng Cheng Laboratory, China} \\
{\textsuperscript{3}Migu Culture Technology Co., Ltd., China}}
\begin{document}
%

\ninept

\makeatletter
\let\@oldmaketitle\@maketitle
\renewcommand{\@maketitle}{\@oldmaketitle
\centering
\vspace{-0.3cm}
\resizebox{0.8\linewidth}{!}{
\includegraphics[trim={1cm 4.5cm 3cm 3.5cm},clip, width=1\linewidth]{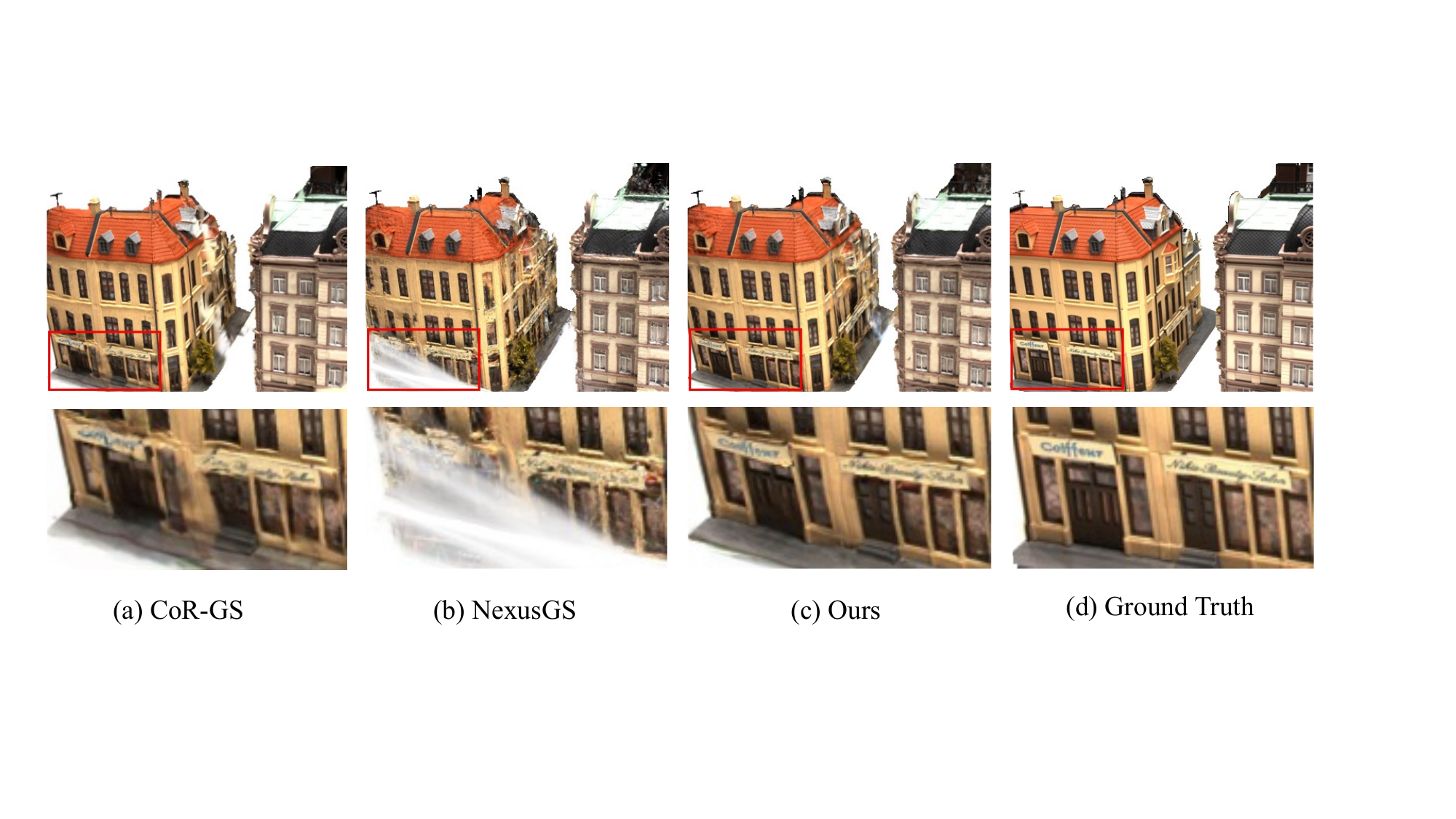}
}

\centering
\captionsetup{hypcap=false}
\captionof{figure}{
\textbf {Qualitative  comparison  of  Sparse-view Novel View Synthesis  quality with the SOTA methods CoR-GS ~\cite{zhang2024cor} and NexusGS ~\cite{zheng2025nexusgs} on DTU ~\cite{jensen2014large} dataset(3 views).} Our method renders more accurate detailed textures.} 
\vspace{10pt}

\label{fig:head}
}

\makeatother

\maketitle
\begin{abstract}

  We present multimodal-prior-guided importance sampling as the central mechanism for hierarchical 3D Gaussian Splatting (3DGS) in sparse-view novel view synthesis. Our sampler fuses complementary cues — photometric rendering residuals, semantic priors, and geometric priors — to produce a robust, local recoverability estimate that directly drives where to inject fine Gaussians. Built around this sampling core, our framework comprises (1) a coarse-to-fine Gaussian representation that encodes global shape with a stable coarse layer and selectively adds fine primitives where the multimodal metric indicates recoverable detail; and (2) a geometric-aware sampling and retention policy that concentrates refinement on geometrically critical and complex regions while protecting newly added primitives in underconstrained areas from premature pruning. By prioritizing regions supported by consistent multimodal evidence rather than raw residuals alone, our method alleviates overfitting texture-induced errors and suppresses noise from appearance inconsistencies. Experiments on diverse sparse-view benchmarks demonstrate state-of-the-art reconstructions, with up to +0.3 dB PSNR on DTU.
  
  \end{abstract}
  
  \section{Introduction}
  \label{sec:intro}
  
  
  Novel view synthesis is a central problem in computer vision with broad applications in virtual/augmented reality, robotics, and content creation. While 3D Gaussian Splatting (3DGS)~\cite{kerbl20233d} provides high-fidelity, real-time rendering with dense multi-view inputs, its performance deteriorates under sparse-view conditions because (i) geometric supervision becomes spatially sparse and uneven, and (ii) the default densification-and-pruning strategy blindly scatters Gaussians, wasting capacity on well-observed surfaces while under-fitting thin structures, object boundaries and texture-rich regions that are essential for photo-realism. Consequently, a key question arises: how can we allocate the limited budget of Gaussians to locations where fine detail is actually recoverable?
  
  Prior efforts to mitigate sparse-view failures include depth-based regularization ~\cite{chung2024depth, li2024dngaussian, zhu2024fsgs}, multi-field regularization  ~\cite{zhang2024cor} , dropout-style regularizers ~\cite{park2025dropgaussian, xu2025dropoutgs}, and approaches that leverage pretrained models to recover dense correspondences or initialize dense point clouds ~\cite{zheng2025nexusgs}. These strategies improve robustness, yet they either impose spatially-uniform constraints or rely on heuristics that do not tell where additional geometry can be reliably recovered.
  


  \begin{figure*}[t]
    \begin{center}
       \includegraphics[trim={2cm 2.5cm 8cm 3cm},clip,width=0.75\linewidth]{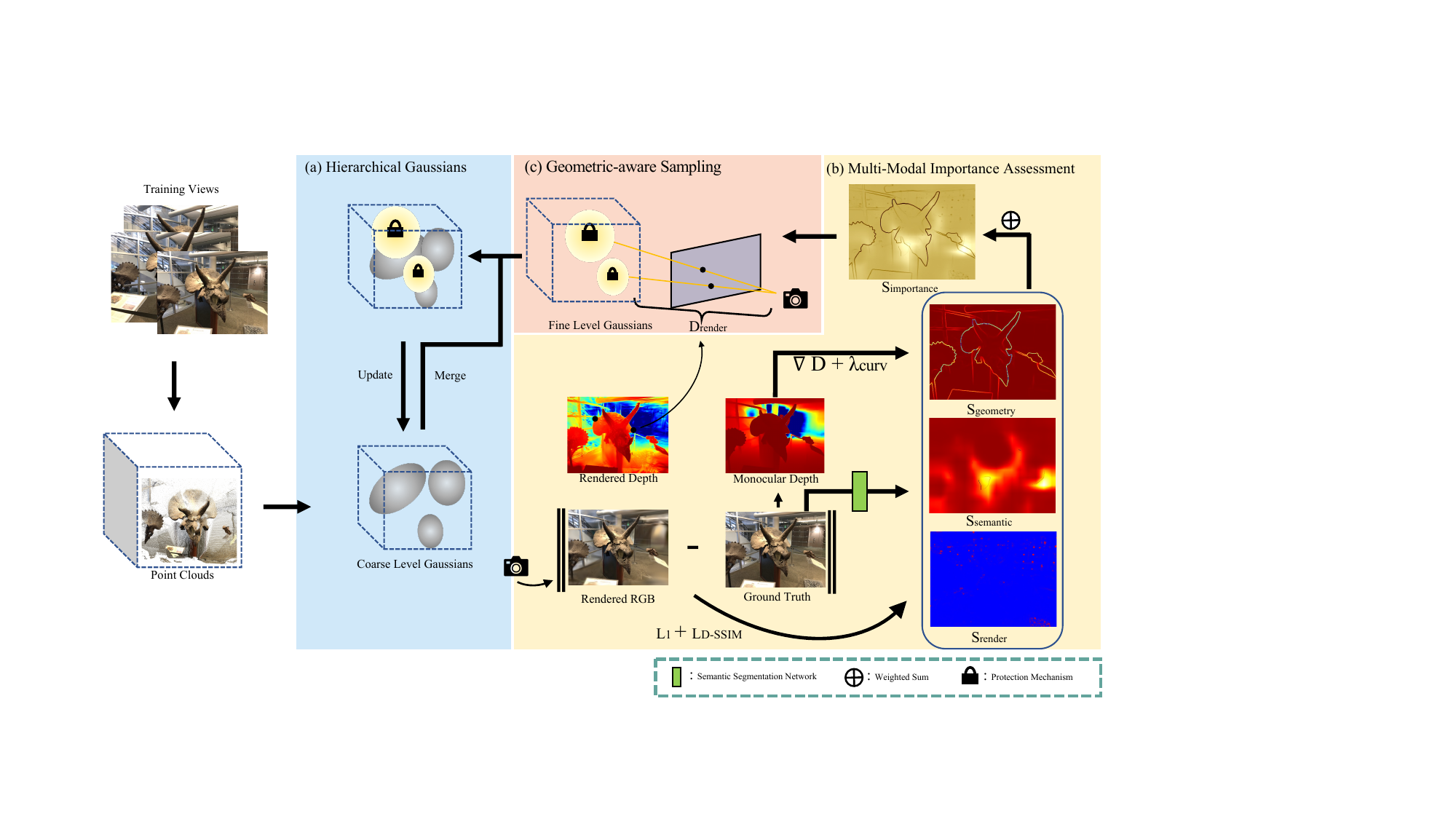}
    \end{center}
    \vspace{-0.3cm}
    \caption{\textbf{Framework.} Our hierarchical Gaussian Splatting framework with Multimodal-Prior-Guided importance sampling for sparse-view novel view synthesis. It consists of three main components: (a) a Hierarchical Gaussian representation with coarse and fine levels (in  \cref{subsec:hier}), (b) a Multi-Modal Importance Assessment module (in \cref{subsec:mmia}), and (c) a Geometric-aware Sampling and pruning strategy (in \cref{subsec:caas}).
     }
    \label{fig:framework}
    \vspace{-0.3cm}
    \end{figure*}

   In contrast,  we propose a hierarchical, geometric-aware 3DGS pipeline driven by multimodal-prior-guided importance sampling. The sampler fuses complementary cues—photometric rendering residuals, semantic priors and geometric priors (e.g., depth and normals) — to produce a local recoverability score that directly determines where to propose fine Gaussians. Built around this sampling core, our framework comprises three key elements:  (1) a multimodal importance metric that discriminates true geometric edges from high-frequency appearance or noise, avoiding the overfitting pitfalls of residual-only strategies; and (2) a coarse-to-fine Gaussian representation in which a stable coarse layer encodes global shape and fine Gaussians are injected selectively where multimodal evidence indicates recoverable detail; and (3) a geometric-aware sampling and retention policy that focuses refinement on geometrically critical and complex regions while protecting newly added primitives in underconstrained areas from premature pruning until sufficient geometric evidence accumulates. Together, these components concentrate modeling capacity where geometry is actually recoverable, preserve sharp boundaries and structural detail, and stabilize optimization under sparse supervision, as shown in \cref{fig:head}.

  We summarize our contributions as follows:
  \begin{itemize}
    \setlength{\itemsep}{2pt}
    \setlength{\parsep}{0pt}
    \setlength{\parskip}{0pt}
  
  \item A multimodal-prior-guided importance metric that fuses photometric, geometric, and semantic signals to localize where fine Gaussians should be allocated.
  \item  A hierarchical 3D Gaussian Splatting framework for sparse-view novel view synthesis that stabilizes optimization via a coarse-to-fine representation driven by multimodal importance estimates.
  \item  A geometric-aware sampling and pruning strategy that concentrates resources on geometrically critical regions and prevents premature removal of newly added primitives in underconstrained areas.
  \end{itemize}
  
  \section{Method}
  \label{sec:method}

  

  We center our design on multimodal-prior-guided importance sampling to address sparse-view novel view synthesis within a hierarchical Gaussian Splatting framework. \cref{fig:framework} depicts the pipeline, whose three main components are: (1) a hierarchical Gaussian representation ( \cref{subsec:hier}) preserves both coarse global shape and selectively added fine detail; (2) a multimodal importance assessment module (\cref{subsec:mmia}) that computes a local recoverability score by fusing photometric residuals with  semantic cues and geometric priors (e.g., depth, normals); and (3) a geometric-aware sampling strategy ( \cref{subsec:caas})that uses the multimodal score to propose, place, and retain fine Gaussians where they are most likely to improve geometry. By making the multimodal sampler the driving mechanism, our method explicitly distinguishes true geometric edges and underconstrained regions from high-frequency appearance or noise, thereby guiding refinement to regions where added primitives yield reliable gains and avoiding wasted or harmful densification.
  
  Given sparse input views $\{I_i\}_{i=1}^N$ with corresponding camera poses $\{P_i\}_{i=1}^N$, our goal is to optimize a 3D scene representation that enables high-quality novel view synthesis. Unlike dense-view scenarios where uniform densification suffices, sparse-view reconstruction must account for spatially varying geometric reliability and selectively allocate modeling capacity accordingly.
  
  \subsection{Preliminaries}
  \label{subsec:prel}
  3D Gaussians Splatting ~\cite{kerbl20233d} is an explicit scene representation that supports high-fidelity real-time renderings. It represents the 3D scene as a set of 3D Gaussians 
  $Gs=\{ G_j| j\in \{1, ..., M \} \}$
  and renders images via splatting: 
  $I^r_{i} =  \Psi(Gs, P_i, K_i),$
  where $I^r_{i}$ is the rendered image and $P_i, K_i$ are the corresponding camera extrinsics and intrinsic.
  Each 3D Gaussian $G_j$ is characterized as $\{ \mu, \Sigma, \alpha, F \}$.
  $\mu$ is the mean of 3D Gaussian distribution. 
  $\Sigma$ is the 3D covariance matrix
  . 
  Besides, $\alpha, F$  are opacity and spherical harmonics coefficients for rendering.
  During rendering, Gaussians are projected to the 2D plane via viewing transformation $W$ and the Jacobian of the affine approximation $J$ of the projection transformation:
  $\Sigma ' = JW \Sigma W^{T} J^{T}.$
  The final color $C$ of each pixel is acquired in an alpha-blending way according to the depth order of the $O$ overlapping Gaussians.
  A tile-based rasterizer is used for efficient rendering.
  During optimization, the Gaussian parameters are updated under the reconstruction loss and SSIM loss ~\cite{wang2004image} between rendered and ground-truth images: $L = (1-\lambda)L_1 + \lambda L_\text{SSIM},$
  where $\lambda$ is the weighting parameter.
  More details can be found in the ~\cite{kerbl20233d}. 
  
  \subsection{Hierarchical Gaussian Representation}
  \label{subsec:hier}
  We introduce a two-level hierarchical structure to balance global shape stability and local detail adaptivity under sparse-view constraints:
  
  \noindent\textbf{Coarse Level Gaussians ($\mathcal{G}_c$):} These Gaussians establish global geometric consistency and provide a stable foundation for the scene structure. They are initialized using the method proposed in ~\cite{zheng2025nexusgs} and remain relatively stable throughout training:$\mathcal{G}_c = \{(\mu_i^c, \Sigma_i^c, \alpha_i^c, c_i^c)\}_{i=1}^{M_c}$
  where $\mu_i^c$, $\Sigma_i^c$, $\alpha_i^c$, and $c_i^c$ represent the position, covariance, opacity, and color of the $i$-th coarse Gaussian, respectively.
  
  \noindent\textbf{Fine Level Gaussians ($\mathcal{G}_f$):} These Gaussians capture detailed geometric features and are adaptively placed based on the propose multimodal importance sampling. They undergo dynamic densification and pruning during training:
  \begin{equation}
  \mathcal{G}_f = \{(\mu_j^f, \Sigma_j^f, \alpha_j^f, c_j^f)\}_{j=1}^{M_f}
  \end{equation}
  
  The final rendering combines contributions from both levels:
  $I_{\text{render}} = R(\mathcal{G}_c \cup \mathcal{G}_f, P)$
  where $R(\cdot, P)$ denotes the Gaussian splatting rendering function for camera pose $P$.
  
  \begin{figure}[t]
    \vspace{-0.3cm}
    \begin{center}
       \includegraphics[trim={4cm 1cm 4cm 0cm},clip,width=0.95\linewidth]{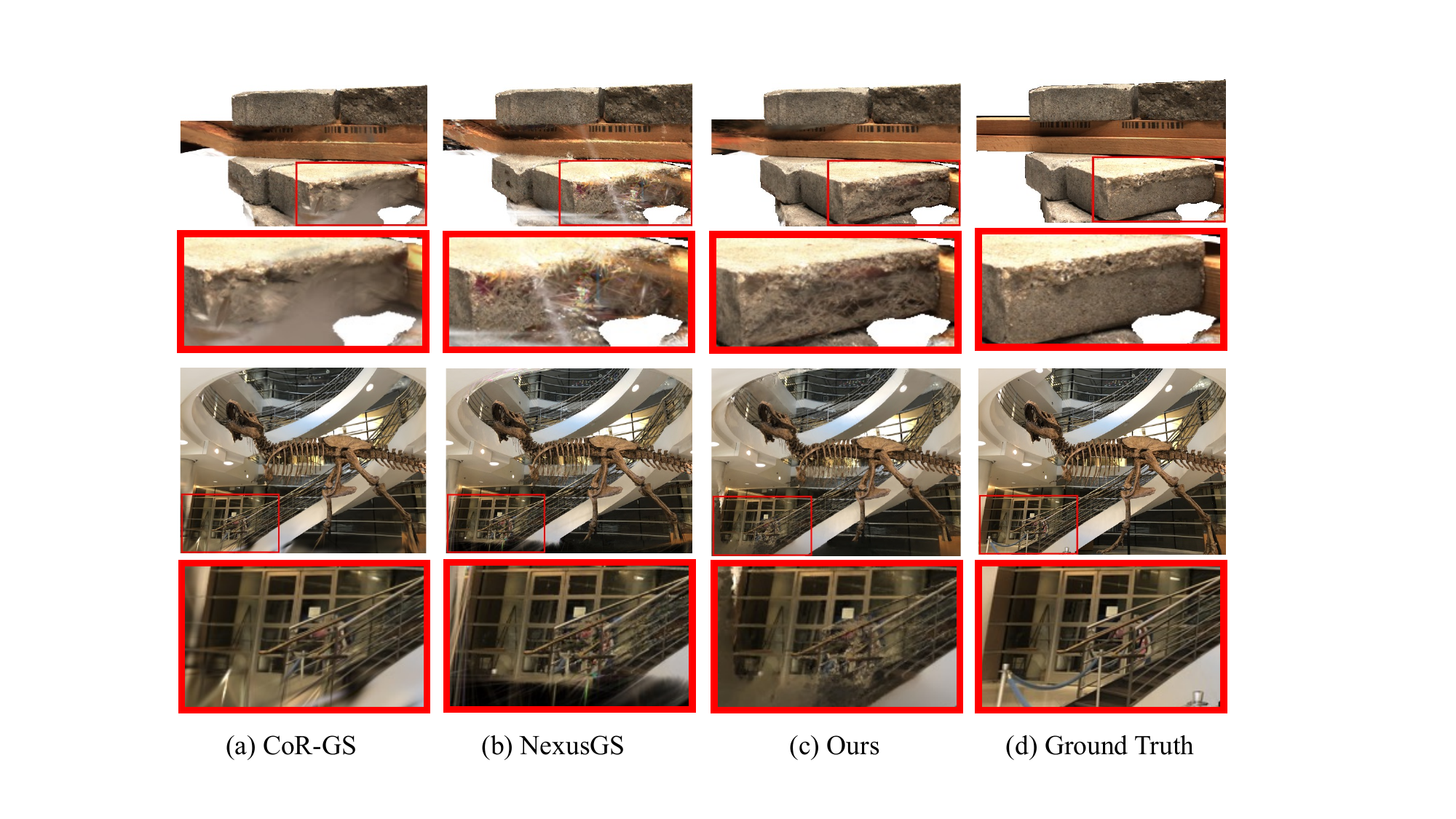}
    \end{center}
    \vspace{-0.3cm}
    \caption{ \textbf{Qualitative results on LLFF ~\cite{mildenhall2019local} and DTU ~\cite{jensen2014large} Datasets.} }
    \label{fig:comparison}
    \vspace{-0.3cm}
    \end{figure}

  \begin{table*}[htbp]
    \scriptsize
    \centering
    \caption{Quantitative comparison on LLFF ~\cite{mildenhall2019local} , DTU ~\cite{jensen2014large}, and MipNeRF-360 ~\cite{barron2022mip} datasets. We color each cell as \colorbox{firstplace} {best}, \colorbox{secondplace} {second best}.}
    \label{tab:quantitative_comparison}
    \resizebox{0.55\textwidth}{!}{
    \begin{tabular}{l|ccc|ccc|ccc}
    \toprule
    \multirow{2}*{Method} & \multicolumn{3}{c|}{LLFF (3 Views)} & \multicolumn{3}{c|}{DTU (3 Views)} & \multicolumn{3}{c}{MipNeRF-360 (24 Views)} \\
    \cmidrule(lr){2-4} \cmidrule(lr){5-7} \cmidrule(lr){8-10}
    & PSNR↑ & SSIM↑ & LPIPS↓  & PSNR↑ & SSIM↑ & LPIPS↓ & PSNR↑ & SSIM↑ & LPIPS↓  \\
    \midrule
    DietNeRF~\cite{jain2021putting} & 14.94 & 0.370  & 0.496 & 11.85 & 0.633 & 0.314 & 20.21 & 0.557 & 0.387  \\
    FreeNeRF~\cite{yang2023freenerf} & 19.63 & 0.612  & 0.308 & 19.92 & 0.787 & 0.182 & 22.78 & 0.689 & 0.323  \\
    SparseNeRF~\cite{wang2023sparsenerf} & 19.86 & 0.624  & 0.328 & 19.55 & 0.769 & 0.201 & 22.85 & 0.693 & 0.315  \\
    \midrule
    3DGS~\cite{kerbl20233d} & 18.54 & 0.588  & 0.272 & 17.65 & 0.816 & 0.146 & 21.71 & 0.672 & 0.248  \\
    DNGaussian~\cite{li2024dngaussian} & 19.12 & 0.591  & 0.294 & 18.91 & 0.790 & 0.176 & 18.06 & 0.423 & 0.584 \\
    FSGS~\cite{zhu2024fsgs} & 20.43 & 0.682  & 0.248 & 17.14 & 0.818 & 0.162 & 23.40 & 0.733 & 0.238  \\
    CoR-GS~\cite{zhang2024cor} & 20.45 & 0.712  & 0.196 & 19.21 & 0.853 & 0.119 & 23.55 & 0.727 & 0.226  \\
    NexusGS~\cite{zheng2025nexusgs} & \colorbox{secondplace}{21.07} & \colorbox{secondplace}{0.738} & \colorbox{secondplace}{0.177} & \colorbox{secondplace}{20.21} & \colorbox{secondplace}{0.869} & \colorbox{firstplace}{0.102} & \colorbox{secondplace} {23.86} & \colorbox{secondplace} {0.753} & \colorbox{firstplace} {0.206}  \\
    \midrule
    \textbf{Ours} & \colorbox{firstplace}{\textbf{21.17}} & \colorbox{firstplace}{\textbf{0.746}} & \colorbox{firstplace}{\textbf{0.175}} & \colorbox{firstplace}{\textbf{20.51}} & \colorbox{firstplace}{\textbf{0.872}} & \colorbox{secondplace}{\textbf{0.104}} & \colorbox{firstplace}{\textbf{23.88}} & \colorbox{firstplace}{\textbf{0.754}} & \colorbox{secondplace}{\textbf{0.208}} \\
    \bottomrule
    \end{tabular}
    }
    \vspace{-0.3cm}
  \end{table*}

  \subsection{Multi-Modal Importance Assessment}
  \label{subsec:mmia}
  To address the limitations of single-criterion sampling, we design a multi-modal importance assessment that integrates three complementary signals:
  
  \noindent\textbf{Rendering Residual ($S_{\text{render}}$):} Measures reconstruction error at each pixel:
  \begin{equation}
  S_{\text{render}}(x,y) = \|I_{\text{gt}}(x,y) - I_{\text{render}}(x,y)\|_2
  \end{equation}
  
  
  \noindent\textbf{Semantic Prior ($S_{\text{semantic}}$):} Leverages a lightweight semantic segmentation network to identify object boundaries and semantically important regions:
  \begin{equation}
  S_{\text{semantic}}(x,y) = \mathcal{F}_{\text{seg}}(I_{\text{ref}})(x,y) \cdot (\omega_{\text{boundary}}(x,y) + \omega_{\text{foreground}}(x,y))
  \end{equation}
  where $\mathcal{F}_{\text{seg}}$ is a ResNet18-based ~\cite{he2016deep}  segmentation network pretrained with 21 semantic classes, $\omega_{\text{boundary}}$ and  $\omega_{\text{foreground}}$ enhances object boundary regions and foreground regions.

  \noindent\textbf{Geometric Complexity ($S_{\text{geometry}}$):} Evaluates local geometric variation with depth gradients:
  \begin{equation}
  S_{\text{geometry}}(x,y) = \|\nabla D(x,y)\|_2 + \lambda_{\text{curv}} \kappa(x,y)
  \end{equation}
  where $D(x,y)$ is the the  monocular depth estimated with the Dense Prediction Transformer (DPT) ~\cite{ranftl2021vision} and $\kappa(x,y)$ represents surface curvature estimated with the second-order gradient of the depth.
  
  
  
  The final importance score combines these signals:$S_{\text{importance}}(x,y) \\ = \mathbf{w}^T \mathbf{s}(x,y)$,
  where $\mathbf{w} = [w_1, w_2, w_3]^T$ are the weighting coefficients and $\mathbf{s}(x,y) = [S_{\text{render}}, S_{\text{semantic}}, S_{\text{geometry}}]^T$ contains the individual importance scores.
  
  \subsection{Geometric-Aware Sampling}
  \label{subsec:caas}
  To ensure robust training, our sampling strategy targets regions with strong geometric constraints while avoiding poorly-constrained areas:

  \noindent\textbf{Reliability Assessment:} To ensure robust training, we identify well-constrained regions:$M_{\text{reliable}}(x,y) = I_g(x,y)$,
    where $I_g(x,y) = \mathbbm{1}[S_{\text{geometry}}(x,y) > \tau_{\text{geometry}}]$ is indicator functions for geometry constraints.
  
  \noindent\textbf{Adaptive Gaussian Placement:} New Gaussians are placed probabilistically based on the importance score, but only in reliable regions:
  $P_{\text{sample}}(x,y) = \frac{S_{\text{importance}}(x,y) \cdot M_{\text{reliable}}(x,y)}{\sum_{(i,j)} S_{\text{importance}}(i,j) \cdot M_{\text{reliable}}(i,j)}$.
  The probabilistic sampling approach prevents over-concentration in high-scoring regions while maintaining exploration of moderately important areas. It avoids local optima in Gaussian placement and ensures better spatial coverage compared to deterministic top-k selection, leading to more robust scene representation. For a sampled pixel $(u,v)$, we back-project it to 3D space to sample fine gaussian using the rendering depth $d$ and camera intrinsics/extrinsics:
  \begin{equation}
    \mathbf{p}_w = R^{-1}\left(d,K^{-1}\begin{bmatrix}u\ v\ 1\end{bmatrix}^{T} - t\right), 
    \end{equation}
  where $K, R, t$ are the camera intrinsic matrix, rotation matrix, and translation vector, respectively. The covariance $\Sigma$ is initialized as an isotropic Gaussian with a predefined scale.
  
  \noindent\textbf{Protection Mechanism:} To prevent premature pruning under sparse supervision, newly added Gaussians are protected for $T_{\text{protect}}$ iterations: $\alpha_{\text{protected}} = \max(\alpha_{\text{original}}, \alpha_{\text{minimum}}) \quad \text{for } t < T_{\text{protect}}$
  This protection mechanism prevents premature removal of newly added Gaussians, which may initially appear suboptimal but possess significant representational potential. By maintaining a minimum opacity threshold during the protection period, we ensure adequate optimization time for new primitives to demonstrate their value, particularly crucial under sparse supervision where individual Gaussians may only contribute meaningfully to a subset of views initially.

  \subsection{Training Strategy}
  \label{subsec:ts}
  Our training procedure alternates between Gaussian optimization and Gaussian sampling:
  
  \noindent\textbf{Phase 1 - Coarse Initialization:} Initialize coarse Gaussians from point clouds and optimize for $N_{\text{coarse}}$ iterations to establish basic scene geometry.
  \textbf{Phase 2 - Hierarchical Refinement:} Iteratively add fine Gaussians based on multimodal importance sampling every $T_{\text{sample}}$ iterations. The sampling frequency decreases over training:
  $T_{\text{sample}}(t) = T_{\text{base}} \cdot (1 + \gamma \cdot t).$
  \textbf{Phase 3 - Stabilization:} In the final training phase, we freeze the Gaussian placement and focus on optimizing parameters for stable convergence.
  
  This hierarchical framework with multimodal importance sampling enables robust sparse-view novel view synthesis by intelligently allocating computational resources where they are most beneficial while maintaining geometric consistency across challenging viewing conditions. During training, we employ the same loss as 3DGS\cite{kerbl20233d}.

    
  
  
  
    
  
  

  \begin{table}[htbp]
    
    \caption{Ablation study on DTU ~\cite{jensen2014large} with 3 training views.} 
    \centering
    \resizebox{0.85\linewidth}{!}{
    \begin{tabular}{cccccccc ccc}
    \toprule
    $Hier$ &$S_{rend}$ & $S_{sem}$ & $S_{geo}$  & $RA$ & $AGP$ & $PM$ & PSNR↑ & SSIM↑ & {LPIPS↓} \\
  
    \midrule
    $\times$  & $\checkmark$ & $\checkmark$    & $\checkmark$    & $\checkmark$ & $\checkmark$    & $\checkmark$ & 20.35 & 0.870  & 0.103 \\

    $\checkmark$  & $\times$ & $\checkmark$    & $\checkmark$    & $\checkmark$ & $\checkmark$    & $\checkmark$ & 20.32 & 0.870  & 0.104 \\
  
    $\checkmark$  & $\checkmark$  & $\times$   & $\checkmark$    & $\checkmark$ & $\checkmark$    & $\checkmark$ & 20.41 & 0.871  & \textbf {0.101} \\
  
    $\checkmark$  & $\checkmark$  & $\checkmark$    & $\times$   & $\checkmark$ & $\checkmark$    & $\checkmark$ & 20.43 & 0.871  & 0.104 \\
  
    $\checkmark$  & $\checkmark$ & $\checkmark$   & $\checkmark$  & $\times$ & $\checkmark$    & $\checkmark$ &  20.30  & 0.869 & 0.103 \\
  
    $\checkmark$  & $\checkmark$  & $\checkmark$    & $\checkmark$    & $\checkmark$ & $\times$     & $\checkmark$ & 20.36 & 0.872  & 0.104 \\
  
    $\checkmark$  & $\checkmark$ & $\checkmark$   & $\checkmark$  & $\checkmark$  & $\checkmark$    & $\times$  &  20.25  & 0.869 & 0.102 \\

    $\checkmark$  & $\checkmark$ & $\checkmark$   & $\checkmark$  & $\checkmark$  & $\checkmark$  & $\checkmark$ & \textbf {20.51}  & \textbf{0.872} & 0.104 \\
    \bottomrule
    \end{tabular}
    }
    \label{tab:ablation}
    \vspace{-0.5cm}
  \end{table}
  
  \section{Experiments}
  \label{sec:exp}
  
  We conduct comprehensive experiments to evaluate our hierarchical 3D Gaussian Splatting framework on sparse-view novel view synthesis tasks. Our evaluation demonstrates superior performance compared to existing methods across multiple datasets and metrics.

  \subsection{Experimental Setup}
  
  \noindent\textbf{Datasets \& Implementation Details.} We evaluate on three standard benchmarks for novel view synthesis: (1) \textbf{LLFF}~\cite{mildenhall2019local} containing 8 forward-facing real scenes, (2) \textbf{DTU}~\cite{jensen2014large} with 15 object-centric scenes under controlled lighting, and (3) \textbf{Mip-NeRF360}~\cite{barron2022mip} synthetic dataset with 9 scenes featuring complex materials and lighting. For sparse-view evaluation, we use 3 training views for DTU and LLFF, and24 views for Mip-NeRF360 following prior work~\cite{zheng2025nexusgs, zhu2024fsgs}.Our framework is implemented in PyTorch with CUDA acceleration. The hierarchical training consists of three phases: coarse initialization (2K iterations), hierarchical refinement (25K iterations for LLFF, 5K iterations for DTU and Mip-NeRF360), and stabilization (3K iterations). Multimodal-prior-guided importance sampling weights are set as $w_1=0.4$, $w_2=0.2$, $w_3=0.4$ based on validation performance.
  
  \noindent\textbf{Baselines \& Evaluation Metrics.} We compare against NeRF methods including: DietNeRF~\cite{jain2021putting}, FreeNeRF~\cite{yang2023freenerf}, SparseNeRF~\cite{wang2023sparsenerf}, and the 3D Gaussian Splatting methods including: 3DGS~\cite{kerbl20233d}, FSGS~\cite{zhu2024fsgs}, DNGaussian~\cite{li2024dngaussian}, CoR-GS~\cite{zhang2024cor} and NexusGS~\cite{zheng2025nexusgs}. All methods are trained using identical sparse view configurations for fair comparison.
  We report standard image quality metrics: PSNR,  SSIM~\cite{wang2004image}, and LPIPS ~\cite{zhang2018unreasonable}.  Higher PSNR and SSIM indicate better quality, while lower LPIPS suggests better perceptual similarity.
  
  \subsection{Quantitative Results}
  \cref{tab:quantitative_comparison} presents quantitative comparisons on all three datasets. Our method consistently outperforms existing approaches across different sparse-view settings.
  On LLFF, our method achieves 21.17 dB PSNR with 3 views, outperforming the best baseline by 0.1 dB. The improvement is even more significant on DTU with 3 views, where we achieve 0.3 dB versus SOTA method NexusGS ~\cite{zheng2025nexusgs}. These results demonstrate the effectiveness of our hierarchical framework in handling severely under-constrained scenarios.
  
  \subsection{Ablation Studies}
  
  We conduct ablation studies to analyze the contribution of each component in our framework, as shown in~\cref{tab:ablation}, where $Hier$ refers to the  hierarchical setting, $S_{rend}$, $S_{sem}$, $S_{geo}$ refers to the three importance metrics in \cref{subsec:mmia}. $RA$, $AGP$, $PM$ refers to the Reliability Assessment, Adaptive Gaussian Placement and Protection Mechanism in \cref{subsec:caas}.
  The results show that all components of the multi-modal importance assessment contribute to the final outcome, and the reliability assessment ensures robust training. Without adaptive Gaussian placement, the Gaussians tend to over-concentrate on certain regions, leading to degraded performance. Likewise, without the protection mechanism, most newly added fine Gaussians would be pruned before they can take effect. The full model achieves the best performance, confirming the complementary benefits of each design choice.
  \subsection{Qualitative Results}
  
  \cref{fig:comparison} presents visual comparisons on challenging scenes. Our method produces sharper details and more consistent geometry compared to baselines, particularly in regions with limited view coverage.
  The improvements are most visible in: (1) fine geometric details like texture patterns, as shown in the first row, (2) reduced artifacts in under-constrained regions, as shown in the second row. These qualitative results align with our quantitative metrics.
  
  \section{Conclusion}
  \label{sec:conc}
  We present a hierarchical framework with multimodal-prior-guided  importance sampling to address sparse-view challenges in 3D Gaussian Splatting. Our method introduces hierarchical gaussians, multimodal importance assessment, and reliability-aware masking to improve geometric supervision and Gaussian placement.
  Experiments demonstrate significant improvements over existing methods.
  The framework improve rendering quality, enabling practical applications in mobile AR/VR and rapid prototyping.
  This work provides a foundation for sparse-view novel view synthesis, demonstrating the effectiveness of integrating multimodal-prior-guided importance sampling with hierarchical gaussians.

\bibliographystyle{IEEEbib}
\bibliography{main}

\end{document}